\documentclass{article}
\usepackage{spconf,amsmath,epsfig}
\usepackage{graphicx}
\usepackage{multirow}
\usepackage{subcaption}
\usepackage{amssymb}

\let\OLDthebibliography\thebibliography
\renewcommand\thebibliography[1]{
  \OLDthebibliography{#1}
  \setlength{\parskip}{0pt}
  \setlength{\itemsep}{0pt plus 0.3ex}
}

\begin{document}\sloppy

\def\x{{\mathbf x}}
\def\L{{\cal L}}

\title{Controllable and Gradual Facial Blemishes Retouching via Physics-Based Modelling}
%
\name{\parbox{\linewidth}{\centering
Chenhao Shuai$^1$, 
Rizhao Cai$^{1\dagger}$\sthanks{Corresponding author}, 
Bandara Dissanayake$^2$, 
Amanda Newman$^2$,\\
Dayan Guan$^1$\sthanks{Rapid-Rich Object Search (ROSE) Lab}, 
Dennis Sng$^{1\dagger}$, 
Ling Li$^{1\dagger}$, 
Alex Kot$^{1\dagger}$\sthanks{This work was carried out at the Rapid-Rich Object Search (ROSE) Lab, School of Electrical \& Electronic Engineering, Nanyang Technological University (NTU), Singapore. The research is supported in part by the A*STAR under it’s A*STAR-P\&G Collaboration (Award H23HW10004), and the NTU-PKU Joint Research Institute (a collaboration between NTU and Peking University that is sponsored by a donation from the Ng Teng Fong Charitable Foundation). Any opinions, findings and conclusions or recommendations expressed in this material are those of the author(s) and do not reflect the views of the A*STAR.}
}}

\address{
$^1$Nanyang Technological University, Singapore\\
$^2$The Procter and Gamble Company, USA
}
\vspace{-3em}

\maketitle

\begin{abstract}
Face retouching aims to remove facial blemishes, such as pigmentation and acne, and still retain fine-grain texture details. Nevertheless, existing methods just remove the blemishes but focus little on realism of the intermediate process, limiting their use more to beautifying facial images on social media rather than being effective tools for simulating changes in facial pigmentation and ance. Motivated by this limitation, we propose our \textbf{C}ontrollable and \textbf{G}radual \textbf{F}ace \textbf{R}etouching (\textbf{CGFR}). Our CGFR is based on physical modelling, adopting Sum-of-Gaussians to approximate skin subsurface scattering in a decomposed melanin and haemoglobin color space. Our CGFR offers a user-friendly control over the facial blemishes, achieving realistic and gradual blemishes retouching. Experimental results based on actual clinical data shows that CGFR can realistically simulate the blemishes' gradual recovering process.

\end{abstract}

\vspace{-1.2em}
\section{Introduction}
\vspace{-0.55em}
\label{sec:intro}
\begin{figure*}[t]
  \centering
  \includegraphics[width=0.885\textwidth]{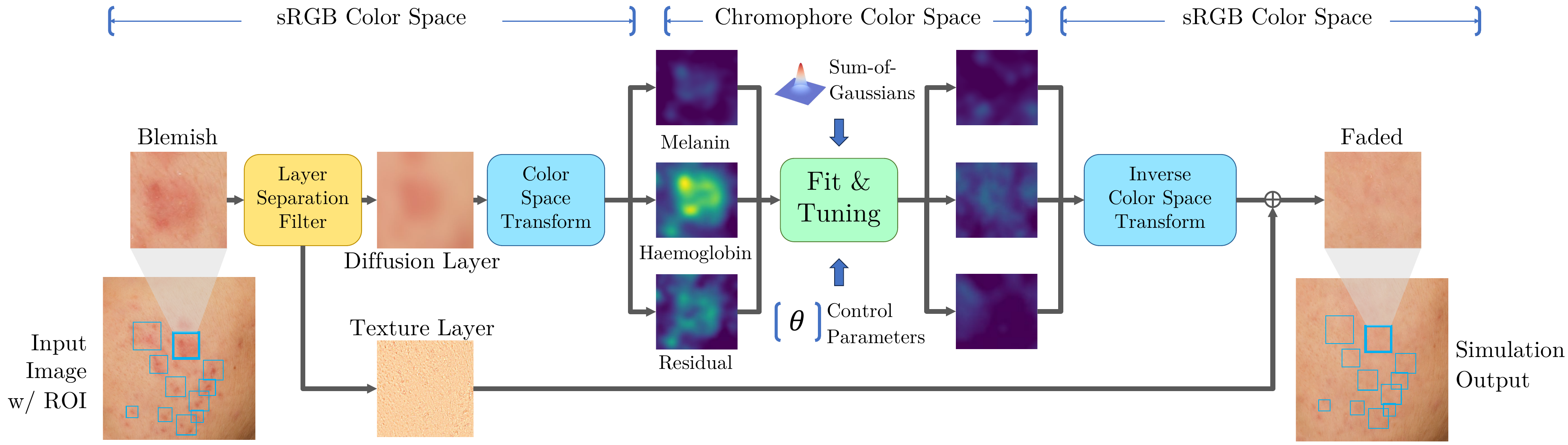}
  \caption{An overview of our CGFR pipeline. In our pipeline, a box of Region of Interest (ROI) is first used to select the blemish like pigmentation or acne. Then, a \textit{Layer Separation Filter} is applied to separate the texture layer and the diffusion layers. A \textit{Sum-of-Gaussians} model is fitted to each ROI in \textit{Melanin/haemoglobin} color space, with the parameters of the fitted model adjusted to manipulate the appearance of the blemishes. The modified diffusion layer is summed with the original texture layer to obtain the output.}
  \label{fig:system}
  \vspace{-0.55em}
\end{figure*}

Face retouching aims at removing facial blemishes while retaining the skin texture details, usually used to beautify facial images to be posted to social media \cite{xieBlemishawareProgressiveFace2023, linExemplarbasedFreckleRetouching2019, shafaeiAutoRetouchAutomaticProfessional2021}. However, current face retouching methods, while adept at completely removing blemishes, lack the capability to represent the gradual and nuanced effects of blemish recovering process due to skincare treatments, which is required to simulate and visualize the efficacy of skincare products \cite{doi:10.2352/EI.2023.35.7.IMAGE-276, doi:10.2352/EI.2022.34.8.IMAGE-300}. 
To bridge the gap between applying face retouching and evaluation of skincare product efficacy, we propose a controllable face retouching that can provide gradual retouching results instead of completely removing the blemishes, named Controllable Gradual Face Retouching (CGFR).

Motivated by the physical properties of skin that the skin blemishes are correlated to the subdermal chromophores (mainly melanin and haemoglobin)\cite{ANDERSON198113}, we suggest converting facial blemish images into chromophore-based color space and modeling their spatial distribution using Sum-of-Gaussians. By adjusting chromophore distribution parameters, we can gradually refine the appearance of skin blemishes. Current retouching methods lack the ability to depict intermediate stages and realistic evolution of blemishes.

To validate our method, we conduct experiments by using our method to gradually retouch facial blemishes to simulate the blemishes' gradual recovery process, and we use our collected clinical data as the ground-truth images to make comparisons. The comparison shows our method can achieve realistic and natural-looking simulation with lower FID scores and fewer artifacts than other algorithms. Moreover, a visual perception study conducted among users of skincare products confirmed the realistic representation of skin blemish changes by our method.

Our approach offers new possibilities in the cosmetic industry, impacting product development and enabling consumers to make informed skincare choices with visualization of product effectiveness. We highlight our contributions as:
\begin{itemize}

  \item We identify the shortcomings of current face retouching methods in the context of skincare product assessment and propose the novel CGFR method; CGFR is a physics-based modelling method instead of a deep learning-based one, which does not rely on paired and large-scale data;
\vspace{-0.55em}
  \item Our experiments demonstrate that CGFR can realistically simulate the gradual recovery of skin blemishes, suggesting that our CGFR is a robust tool for skin science research;
\vspace{-0.55em}
    \item Our research provides a new use case for the application of computer vision algorithms in the cosmetic and skincare industry and offers promising prospects in product development.
\vspace{-0.55em}
\end{itemize}
\begin{figure}[t]
  \centering
  \vspace{-0.5em}
  \includegraphics[width=0.75\columnwidth]{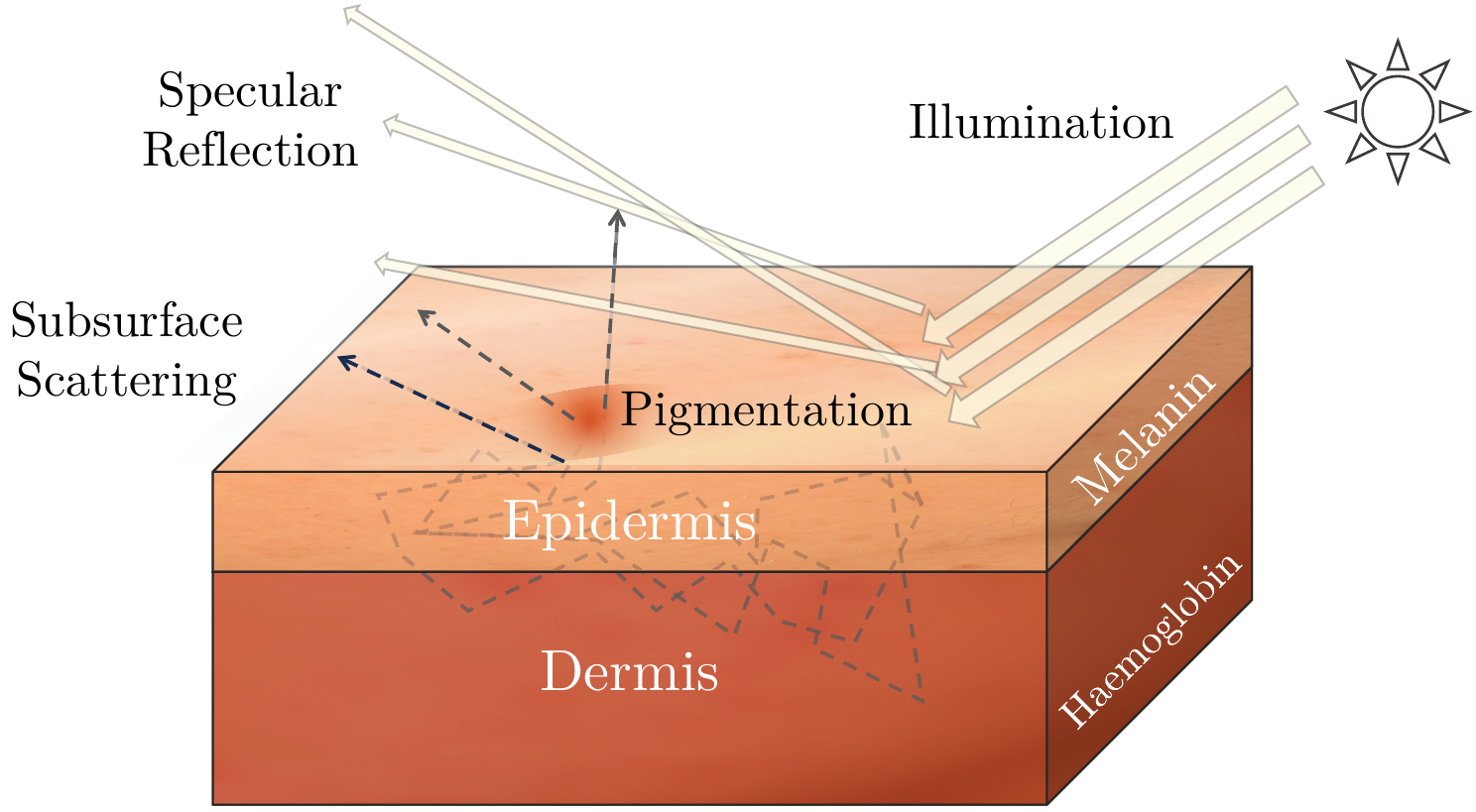}
  \vspace{-0.55em}
  \caption{Layered skin model. A portion of the incident light undergoes specular reflection, revealed as a skin texture layer. The other part transmits into and is scattered by the Epidermis and Dermis. Melanin and haemoglobin, which are distributed in these two layers, absorb specific wavelengths of light, rendering the skin's characteristic color.}
  \label{fig:skin_model}
  \vspace{-1.2em}
\end{figure}
\vspace{-0.65em}
\section{Related Work}
\vspace{-0.5em}
\subsection{Facical Image Retouching}
\vspace{-0.6em}
The most related task to our work is facial image retouching, which focuses on removing blemishes while preserving skin details in facial images as much as possible. We categorize current face retouching and editing methods into three types: "Pixel Space," "Latent Space," and "Parametric Space":

\vspace{-0.7em}
\subsubsection{Pixel Space}
This category includes methods that directly work on \textit{pixel level}, such as inpainting and filtering. Inpainting algorithms utilize neighbouring or similar pixels to fill in blemish positions \cite{lipowezkyAutomaticFrecklesDetection2008, Liu2017AutomaticFF}. However, a simple interpolation between modified and original images and a hard threshold to detect blemish pixels do not conform to the real progression of blemish changes, leading to unnatural editing traces. Filtering methods smooth out blemishes pixels in a content-aware manner to retain skin details while removing unwanted skin blemishes\cite{velusamyFabSoftenFaceBeautification2020}. Although the intensity of filtering is controllable, these methods do not account for the intermediate process of blemish fading.

\vspace{-0.7em}
\subsubsection{Latent Space}
Deep generative models aim to learn a mapping from latent noise to pixels\cite{rombach2021highresolution}. Once the model is trained, modifying the input noise will affect the features of the generated images. We consider these methods as \textit{latent space} editing. Recent works have been explored for skin pigmentation generation\cite{beharaSkinLesionSynthesis2023} but lack explicit control over the output nor the ability to modify existing images. On the other hand, works like \cite{xieBlemishawareProgressiveFace2023, liuAutomaticBeautificationGroupPhoto2018} use GANs for face image retouching. Although latent space interpolation in these models allows for smooth transitions between images, the lack of data annotation means users can hardly control the image changes explicitly.

\vspace{-0.6em}
\subsubsection{Parametric Space}
Physics-based methods explicitly model properties like skin optics and physiology, allowing facial image modification by adjusting model parameters. We define these methods as \textit{Parametric Space} editing. Previous physics-based models predominantly address full face color changes by adjusting chromophore levels, as elaborated in \cite{jungDeepLearningbasedOptical2023a,tsumuraImagebasedSkinColor}. Recently, Lin et al.\cite{linExemplarbasedFreckleRetouching2019} proposed an approach for modifying chromophore content in pigmentation, but it involves a holistic adjustment, which detects pigmentation area pixels and performs histogram mapping to align these pixels to normal skin in color.

Our method is a novel physical-based model, enabling per-spot, per-chromophore concentration modelling and retouching without the need for paired image datasets. It decomposes colors in logarithmic RGB space, more accurately representing the absorption of light by chromophores. Rather than classifying pixels based on a threshold, our approach effectively considers blurred edges of blemishes due to subsurface scattering, allowing for seamless integration of adjusted blemishes with the surrounding skin. To the best of our knowledge, this is the first method that offers precise control over modifications in blemishes caused by localized accumulation of chromophores.


\vspace{-0.55em}
\subsection{Layered Skin Model}
Modelling skin as a layered, semi-transparent material has become common practice in skin rendering and facial image retouching\cite{10.5555/2383894.2383946, luFacialSkinBeautification2016}. We summarize the layered model of skin and their interactions with light as follows:
\begin{enumerate}
  \item \textbf{Specular reflection:} Light reflection from the surface, caused by oils, water, and stratum corneum of the skin. It captures the surface texture of the skin, such as fine grooves and textures. In CGFR, we call it \textit{Texture Layer} and we preserve it unaltered, as shown in Fig.\ref{fig:system}.
        \vspace{-0.55em}
  \item \textbf{Subsurface scattering and absorption:} Skin is semi-transparent\cite{Igarashi2005TheAO}, and its components like the extra-cellular matrix cause incoming light rays to scatter. Some light rays reflect back to the surface, a phenomenon known as subsurface scattering. Local accumulation of skin chromophores results in visible blemishes with distinct colors from the surrounding skin\cite{ANDERSON198113}. On this basis, Jensen et al.\cite{10.1145/3596711.3596747} introduced the Bidirectional Surface Scattering Reflectance Distribution Function (BSSRDF) to approximate scattering media for realistic skin rendering. Jensen et al.\cite{10.1145/1073204.1073308} later used a sum-of-Gaussian function to effectively approximate the scattering of multi-layered skin. Our CGFR focuses on this optical phenomenon for realistic blemish modelling and editing. In our method, we name this layer as the \textit{Diffusion Layer} in Fig.\ref{fig:system} and we propose to use the sum of multiple Gaussian functions to model the blurred appearance of blemish under subsurface scattering in the Diffusion Layer.

\vspace{-0.55em}
  \item \textbf{Transmission:} When the light is very strong and shines on thin tissue (such as the ears or fingers under strong light), a unique transmission appearance can be observed against the light source. In CGFR, we disregard this.
\end{enumerate}
\begin{figure}[t]
  \centering
  \includegraphics[width=0.7\columnwidth]{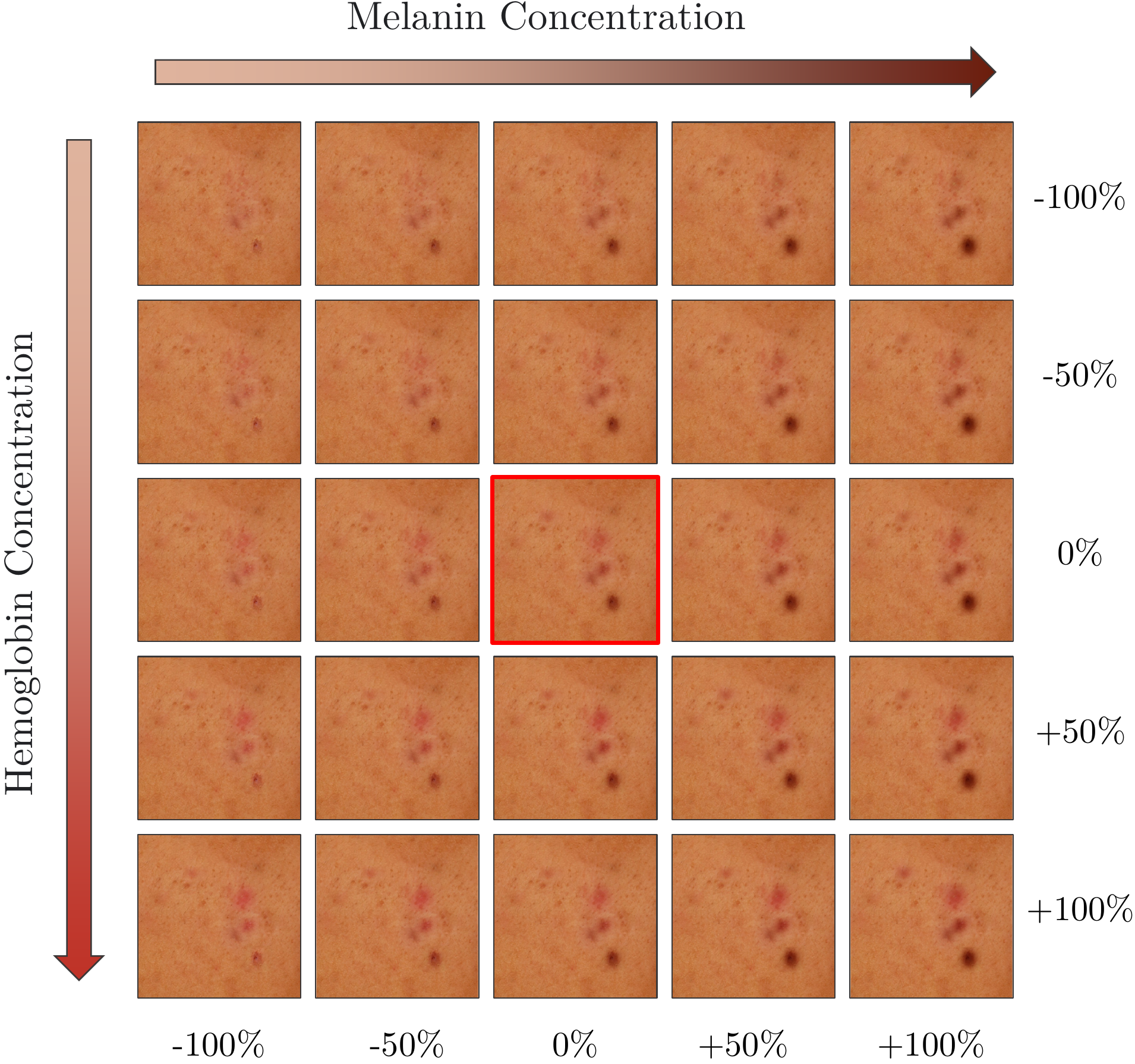}
  \vspace{-0.4em}
  \caption{Matrix of different chromophore concentrations setting. We set different gains $\alpha$ according to equation \ref{eqn:alpha}, shown as percentage values in the figure. The original image is marked by a red box. Our model fully decouples the major chromophores of human skin, enabling highly controllable blemish editing.}
  \label{fig:matrix}
\end{figure}
\vspace{-2.2em}
\section{Methods}

\vspace{-0.55em}
\subsection{Skin Chromophore Color Space Decomposition}
Skin color in digital photos represents a specific subset of the sRGB space, influenced predominantly by chromophores like melanin and haemoglobin\cite{10.1145/1073204.1073308}. Specifically, we model the \textit{relative} concentration of chromophores in a blemished area compared to the surrounding normal skin. According to the Beer-Lambert law, the higher the relative concentration, the more light is absorbed:
\vspace{-0.5em}
\begin{equation}
  A(\lambda) = -log(R(\lambda)) = C\epsilon l,
  \label{eq1}
\end{equation}
where $A$ represents absorption, $R$ is the reflection intensity, $\lambda$ is the wavelengths, $C$ is the relative concentration, $\epsilon$ denotes the extinction coefficient of chromophore and $l$ is the mean optical path length. In our work, we mainly consider the impact of melanin, haemoglobin, and a residual term, as shown in Fig.\ref{fig:skin_model}. Therefore, $C\epsilon l$ in equation \ref{eq1} expands as
\begin{equation}
  A(\lambda) = C_H\epsilon_H(\lambda)l_H + C_M\epsilon_M(\lambda)l_M + C_r\epsilon_r(\lambda)l_r,
  \label{eq2}
\end{equation}
where subscript $H$, $M$, and $r$ represent heamoglobin, melanin and residual chromophore, respectively.

In our method, as shown in Fig.\ref{fig:system}, we are mainly concern about the Diffusion Layer, so we separate it from the input image by a Gaussian filter. We then apply inverse gamma correction to the sRGB image to obtain the linear RGB value, followed by taking its logarithm as the approximation of real $R$. Although it may not fully reflect the real case, it is sufficient to estimate the relative concentration ratio against the surrounding skin. Considering the response of each chromophore under the three camera pixel channels R, G, and B, Eq.\ref{eq1} and Eq.\ref{eq2} can be written as:
\begin{equation}
  \begin{aligned}
     & C_H\epsilon_H^c l_H + C_M\epsilon_M^c l_M + C_r\epsilon_r^c l_r = -log(R^c) \\
     & c\in\{\mathcal{R},\mathcal{G},\mathcal{B}\},
  \end{aligned}
\end{equation}
or in matrix form
\vspace{-0.5em}
\begin{gather*}
  \mathbf{E}\mathbf{c}=-log(\mathbf{k}),\\
  \mathbf{E}=\begin{bmatrix}
    \epsilon_H^\mathcal{R} l_H & \epsilon_M^\mathcal{R} l_M & \epsilon_r^\mathcal{R} l_r \\
    \epsilon_H^\mathcal{G} l_H & \epsilon_M^\mathcal{G} l_M & \epsilon_r^\mathcal{G} l_r \\
    \epsilon_H^\mathcal{B} l_H & \epsilon_M^\mathcal{B} l_M & \epsilon_r^\mathcal{B} l_r
  \end{bmatrix},\
  \mathbf{c}=\begin{bmatrix}C_H \\C_M \\C_r\end{bmatrix},\
  \mathbf{k}=\begin{bmatrix}R^\mathcal{R} \\R^\mathcal{G} \\R^\mathcal{B}\end{bmatrix}.
\end{gather*}

It is reported that there is no significant difference in skin thickness between human races\cite{Whitmore2000}, so we combine $l$ with $\epsilon$ and estimate $\mathbf{E}$ by Fast Independent Component Analysis (FastICA)\cite{HYVARINEN2000411}. 
\vspace{-0.55em}
\subsection{Controllable Blemish Retouching}
Given a 3-channel image patch $C(\mathbf{x})\in\mathbb{R}^{3\times h\times w}, \mathbf{x}\in\mathbb{R}^2$ with length $w$ and height $h$, containing a user-selected blemish, CGFR first uses a Gaussian low-pass filter $g(\cdot)$ to separate the skin texture $C^{text}$ and base color $C^{base}$ parts:
\begin{equation}
C^{base} = g(C(\mathbf{x})), \quad C^{text} = C(\mathbf{x})-C^{base}.
\end{equation}
For $C^{base}$, CGFR assumes it can be divided into normal skin $C_K^{skin}$ and blemish $C_K^{blem}$ in chromophore color space as
\begin{equation}
  C_K^{base}(\mathbf{x}) = C_K^{skin}(\mathbf{x}) + C_K^{blem}(\mathbf{x}),\quad K\in\{H,M,r\}.
\end{equation}
For $C_K^{skin}$, CGFR assumes that skin color changes are smooth within a local area and thus fits it by a simple linear model:
\begin{equation}
  \hat{C_K^{skin}}(\mathbf{x};\mathbf{k},d)=\mathbf{k}\mathbf{x}+d,\quad\mathbf{k}\in\mathbb{R}^2, d\in\mathbb{R}.
\end{equation}
For $C_K^{blem}$, CGFR is based on the observation that pigmentation and acne of interest tend to have blurred edges due to the subsurface scattering. Thus, CGFR adopts 2D Gaussian functions $G$ to describe this effect, in which $G$ is defined as
\begin{equation}
\vspace{-0.2em}
  G(\mathbf{x}; a, \mathbf{\mu}, \Sigma)=\frac{a}{2\pi\sqrt{|\Sigma|}}e^{-\frac{1}{2}(\mathbf{x}-\mathbf{\mu})^T\Sigma^{-1}(\mathbf{x}-\mathbf{\mu})},
\vspace{-0.2em}
\end{equation}
where $\mathbf{\mu}=[\mu_x,\mu_y]^T$ is the centre coordinate $(x,y)$ on the input image plane, $\Sigma\in\mathbb{R}^{2\times2}$ is the covariance matrix and $a$ is the amplitude factor. Second, to better fit variously shaped blemishes on uneven facial areas, CGFR decomposes $\Sigma$ into rotation matrix $\mathbf{R}$ and scaling matrix $\mathbf{S}$ as $\Sigma= \mathbf{R}\mathbf{S}\mathbf{S}^T\mathbf{R}^T$. This allows $G$ to stretch and off-axis rotation with an angle of $\theta\in[0,\pi)$, forming ellipsoidal patterns, that is
\begin{equation}
\vspace{-0.1em}
    \mathbf{R} = \begin{bmatrix}
    \cos\theta&-\sin\theta\\
    \sin\theta&\cos\theta
    \end{bmatrix},\quad\mathbf{S}=diag(\sigma_x,\sigma_y).
\vspace{-0.1em}
\end{equation}
Finally, $\hat{C_K^{blem}}(\mathbf{x})$ is modelled and estimated by sum of $N$ Gaussian functions:
\vspace{-0.5em}
\begin{equation}
\begin{aligned}
    &\hat{C_K^{blem}}(\mathbf{x}; \mathbf{\Theta})=\sum_{i=0}^{N-1}G_i(\mathbf{x}; \Theta_i),\\&\ N\ge1, \Theta_i=\left[a_i, \mathbf{S}_i, \mathbf{\mu}_i, \theta_i\right].
\end{aligned}
\vspace{-0.2em}
\end{equation}
CGFR fits $\hat{C^{base}_K(\mathbf{x})} = \hat{C_K^{blem}}(\mathbf{x}) + \hat{C_K^{skin}}(\mathbf{x})$ to the input by least-squares\cite{10.1007/BFb0067700}. In each sub-step, $G_N$ is gradually added to fit the residual from the previous $N-1$ step, with the existing parameters kept fixed. After all $G_i$ are integrated, a final fine-tuning fitting with all parameters unfrozen is conducted. After the fitting has converged, $\hat{C_K^{skin}}$ is discarded. Instead, $\hat{C_K^{blem}}$ is multiplied with user-input gain $\alpha_K\in\mathbb{R}$ and added with the input to amplify/attenuate the blemish intensity to obtain the modified image patch $C_K^{\prime}$, namely
\begin{equation}
    C_K^{\prime} = C_K^{text} + C^{base}_K+\alpha_K\hat{C_K^{blem}}.
    \label{eqn:alpha}
\end{equation}
Obviously, if $\alpha_K=-1$, the facial blemish will be completely removed from the image patch, while a positive gain will intensify this blemish.
\begin{figure}[t]
  \centering
  \begin{subfigure}{0.35\textwidth}
    \centering
    \includegraphics[width=\linewidth]{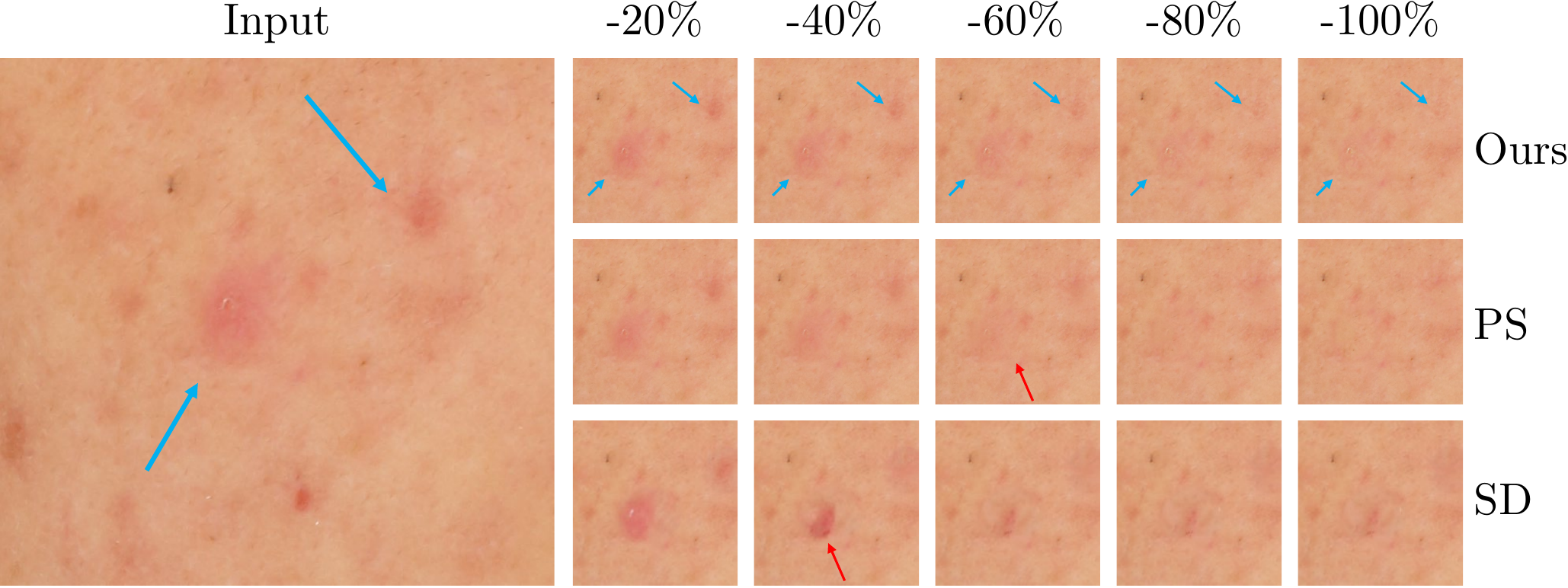}
  \end{subfigure}
  \hfill
  \begin{subfigure}{0.35\textwidth}
    \centering
    \includegraphics[width=\linewidth]{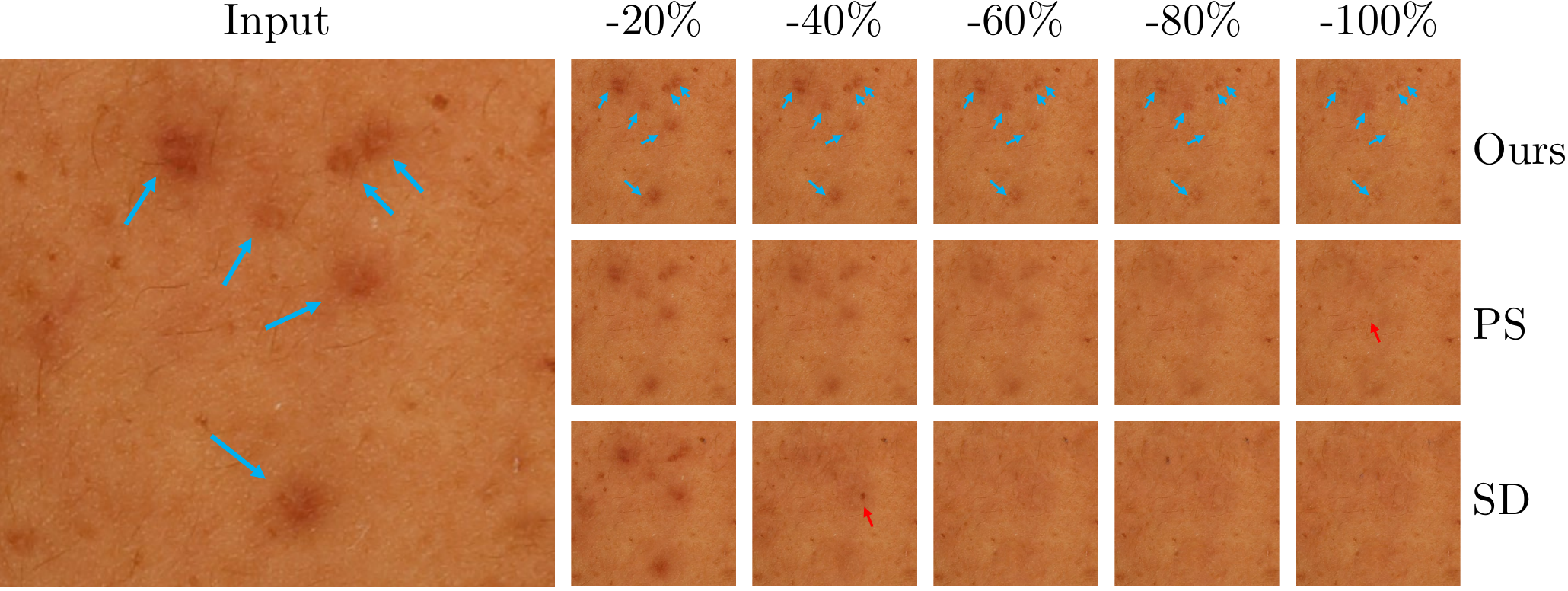}
  \end{subfigure}
  \vspace{-0.55em}
  \caption{Comparison with baseline methods. We compared the results of several blemish removal or modification methods, including our method (marked as Ours), Adobe Photoshop\cite{adobephotoshop} inpainting (marked as PS), and Stable Diffusion\cite{rombach2021highresolution} inpainting (marked as SD). Arrows are manually added to highlight user-selected blemishes. Note the red arrows where the PS produces over-smoothed skin patches and the SD produces visible artifacts.}
  \label{fig:baseline}
  \vspace{-0.55em}
\end{figure}
\begin{figure*}[h]
  \centering
  \begin{subfigure}{0.27\textwidth}
    \includegraphics[width=0.95\linewidth]{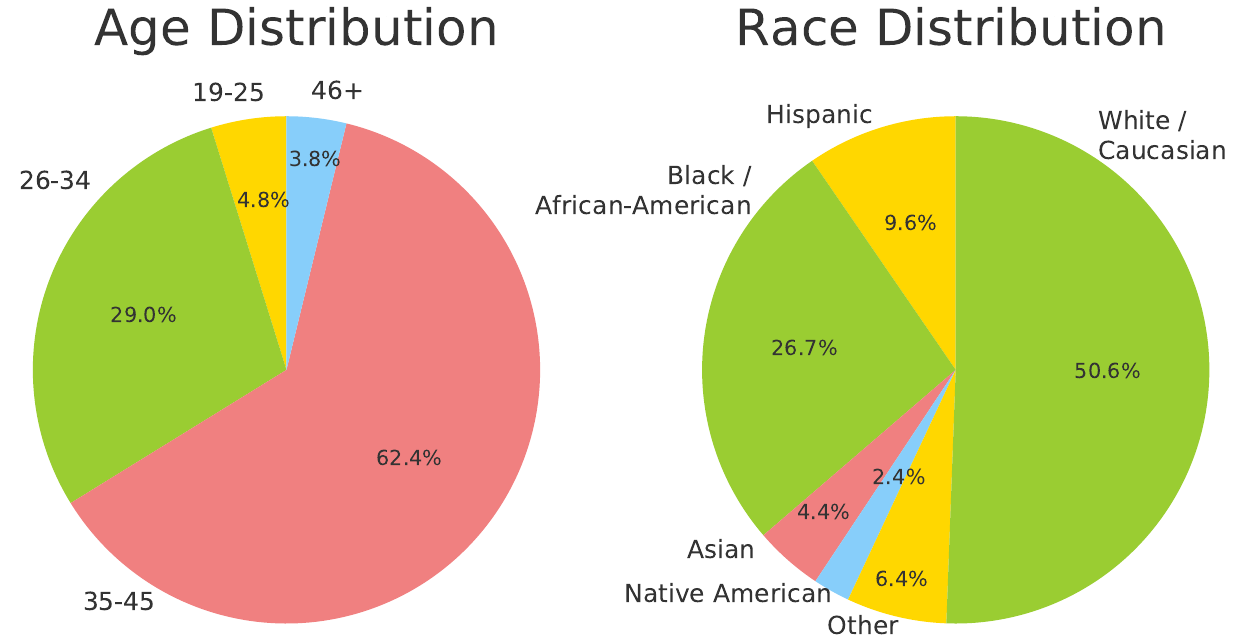}
    \caption{Metadata of panellists}
    \label{fig:metadata}
  \end{subfigure}\hfill
  \begin{subfigure}{0.31\textwidth}
    \includegraphics[width=0.93\linewidth]{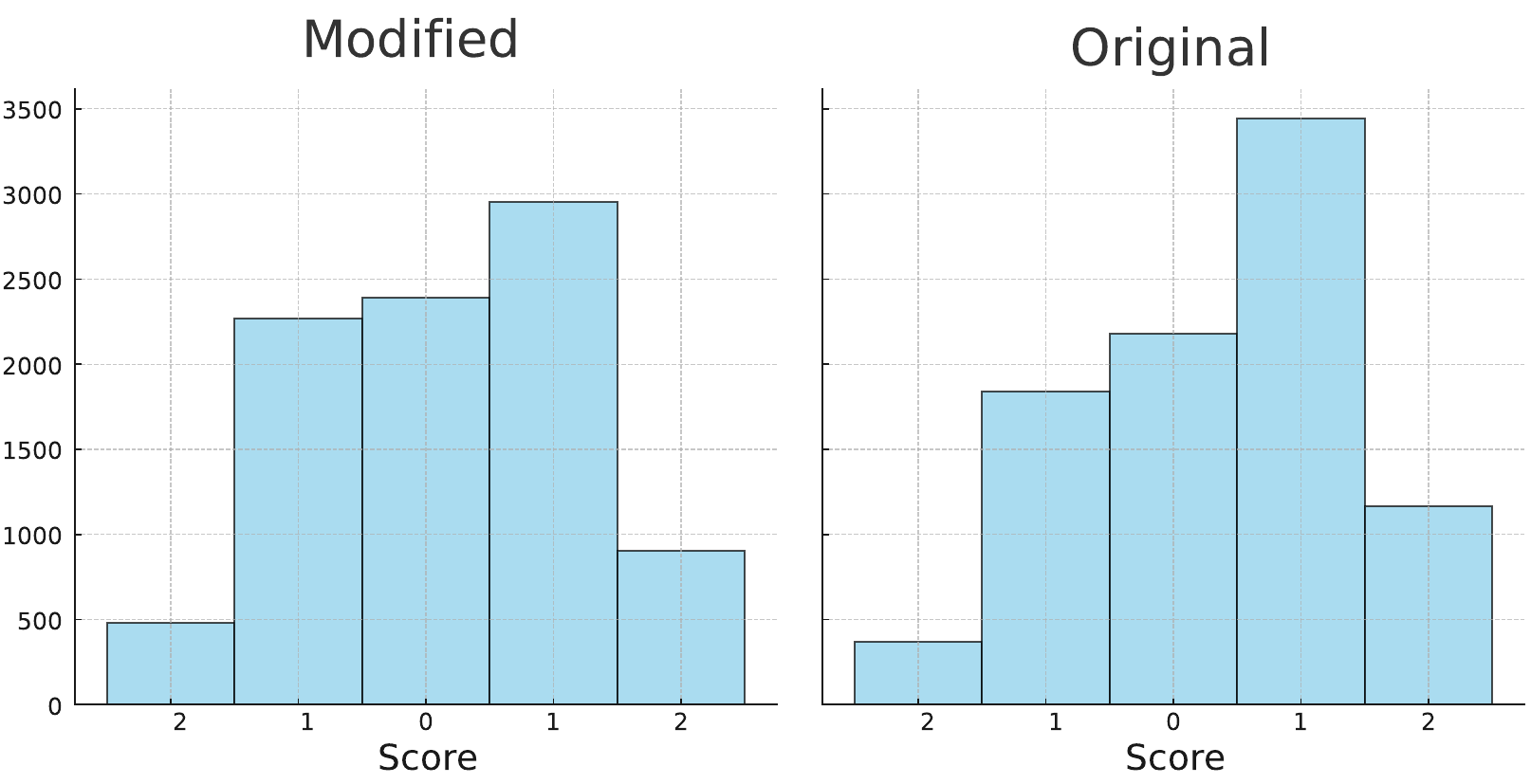}
    \caption{Scoring frequencies of survey responses}
    \label{fig:survey_hist}
  \end{subfigure}\hfill
  \begin{subfigure}{0.34\textwidth}
    \includegraphics[width=0.95\linewidth]{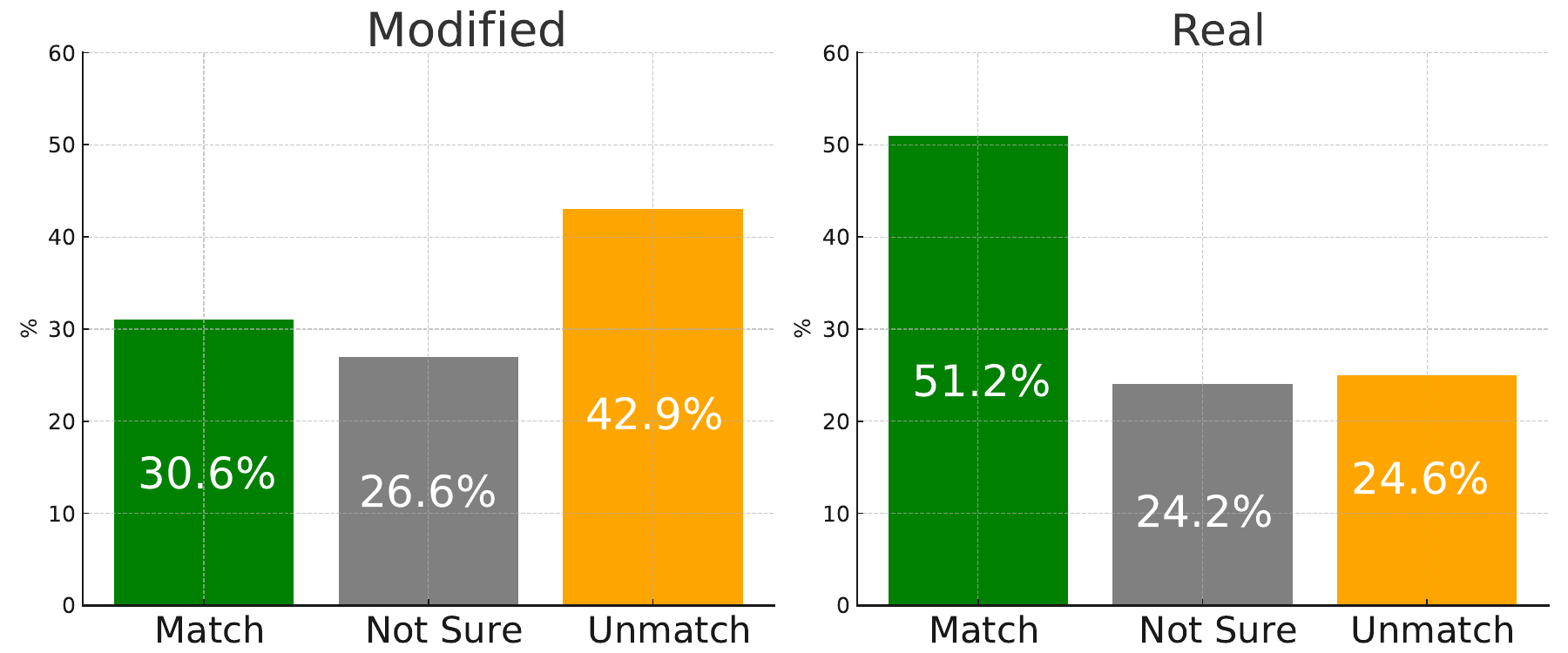}
    \caption{Bar chart of survey results}
    \label{fig:bar_charts}
  \end{subfigure}
  \caption{Metadata and results of the perception study. The test population covers people from 19 to 45 years old, multiple races, and multiple skin tones, as shown in Fig.\ref{fig:metadata}. Panellists scored images from -2 to +2 to assess their confidence in considering the image as modified or not, with higher scores indicating that the user considered the image to be unmodified. Scoring frequencies are displayed in Fig.\ref{fig:survey_hist}. The results of the survey are shown in Fig.\ref{fig:bar_charts}. For the modified images, more people perceived them as unmodified or not sure. This suggests that our modifications are consistent with human perception and intuition.}
  \vspace{-1em}
\end{figure*}
\begin{figure}[t]
  \centering
  \includegraphics[width=0.72\columnwidth]{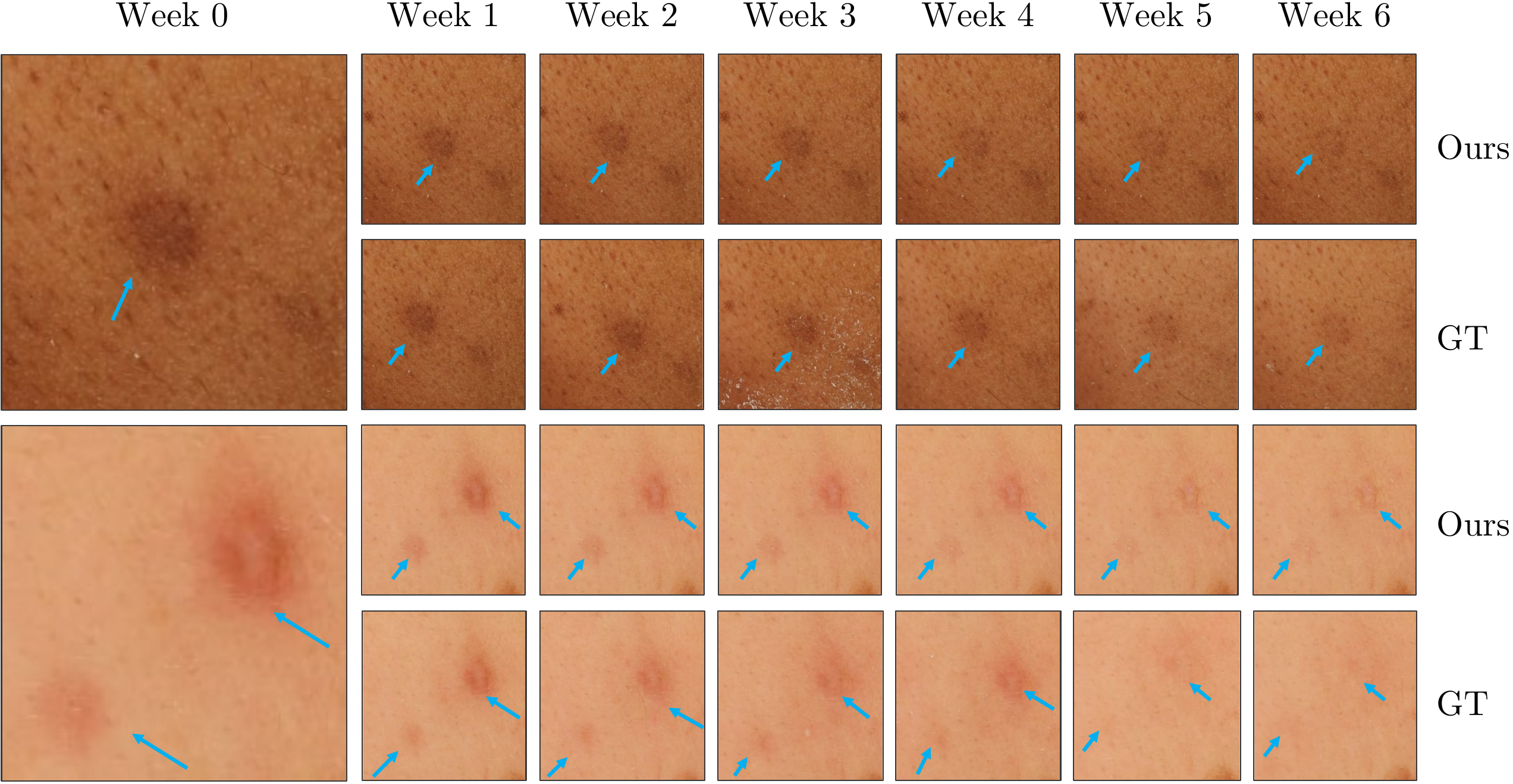}
  \caption{We show the application of our method to the simulation of the fading process of skin blemish. Blue Arrows are manually added to highlight blemishes of interest (user-selected acne or pigmentation). In our simulation, we input images of \textit{Week 0} and adjust $\alpha$ in Eq.\ref{eqn:alpha} of the obtained model to simulate the change of the blemishes in the following weeks. Note that our method applies to different skin tones and various types of blemishes.}
  \label{fig:forward}
  \vspace{-1.4em}
\end{figure}
\vspace{-0.3em}
\section{Experiment \& Results}
\vspace{-0.3em}
\subsection{Dataset}
We adopted a self-collected dataset for our research, development, and testing. The dataset collection is done by two clinical imaging systems (Visia CR4 and OLE both developed by Canfield Scientific), which is spanned over a period of 12 weeks such that the recovering processes of facial blemishes are recorded for us to make comparisons to the results of our CGFR.

\vspace{-0.4em}
\subsection{Experiment Setup}
In this study, we have carried out extensive blemish change simulation experiments to evaluate the algorithm's effectiveness, and evaluated simulation quality in terms of controllability, reality, and versatility. A detailed discussion of each aspect follows.
\begin{table}[t]
  \caption{FID scores of different blemish fading rates. Lower scores are better.}
  \centering
  \resizebox{0.8\columnwidth}{!}{%
    \begin{tabular}{cccccc}
      \hline
      \multirow{2}{*}{Methods} & \multicolumn{5}{c}{Fading Rate}                                                                         \\ \cline{2-6}
                               & 100\%                           & 80\%            & 60\%            & 40\%            & 20\%            \\ \hline\hline
      SD                       & 144.89                          & 133.53          & 134.16          & 160.10          & 159.54          \\
      PS                       & 117.98                          & 120.37          & 125.15          & 129.26          & \textbf{129.96} \\
      \textbf{Ours}            & \textbf{115.30}                 & \textbf{118.12} & \textbf{122.64} & \textbf{127.09} & 131.60          \\ \hline
    \end{tabular}%
  }
  \vspace{-1em}
  \label{tbl:fid}
\end{table}

\vspace{-0.7em}
\subsubsection{Controllability}
One significant advantage of our model is its high controllability, allowing users to freely adjust the parameters of the blemishes to precisely control their appearance.
We plotted a changing matrix by adjusting the concentration control parameters $\alpha$ of melanin and haemoglobin, shown in Fig.\ref{fig:matrix}. We also used CGFR to replicate the natural fading of skin blemishes over time, as illustrated in Fig.\ref{fig:forward}. Our method successfully decouples the concentrations of these two chromophores, allowing users to independently control their appearances, thus flexibly simulating the change of blemishes.

\vspace{-1.6em}
\subsubsection{Reality}
\vspace{-0.3em}
Our algorithm was tested in real human skin conditions. We select widely available and easily reproducible baseline methods for comparison. For \textit{pixel space} editing methods, we choose Adobe Photoshop (PS), the most commonly used image editing and retouching software. For \textit{latent space editing} methods, Stable Diffusion (SD)\footnote{https://huggingface.co/runwayml/stable-diffusion-v1-5}, recognised for its high-quality generation, is our selection.


We assessed the quality of skin blemish editing using the Fréchet Inception Distance (FID) scores\cite{heusel2017gans}, which are displayed in Table \ref{tbl:fid} and visually in Fig.\ref{fig:baseline}. Our method generally achieved lower FID scores, indicating better performance in most cases, except for a 20\% fading rate. Visually, our algorithm outperformed others, maintaining skin details more effectively. PS method resulted in blurred patches, while SD produced some coherent details but with noticeable artifacts and color mismatches at higher denoising ratios.
\vspace{-0.7em}
\subsubsection{Versatility}
The algorithm's versatility enables its generalization across different scenarios. We tested it on various skin tones and blemish types like pigmentation and acne, demonstrating its effectiveness. The algorithm accurately simulates skin blemish changes, preserving subtle skin textures like fine hairs and pores, evident in Fig.\ref{fig:baseline}, highlighting its versatility.
\vspace{-0.7em}
\subsection{Perception Study}
We evaluated the performance of our CGFR by a visual perception study. The aim was to comprehensively evaluate whether the algorithm could produce authentic and believable blemish changes and to analyse whether there are biases in certain attributes of the skin, such as skin color or age.

\vspace{-0.7em}
\subsubsection{Result}
In the perception study, 48 images (24 simulated through our algorithm and 24 unaltered images) were shown to the panellists one image at a time. When the score ranged from 0 to +2, we considered the respondents to be affirming the image as ``real" rather than modified. In the test, as shown in Fig.\ref{fig:bar_charts}, the altered images had a lower average score (0.16956 vs 0.35511), only 30.6\% correctly identifying the altered image vs 23.6\% judging the real images as altered, and 26.6\% not sure if it is or is not altered. This indicates that the effect of our proposed algorithm is superior, to the point where laypeople cannot readily discern traces of alteration.

\vspace{-0.5em}
\section{Conclusion}
\vspace{-0.5em}
We introduced CGFR, a novel method combining a physics-based model with dermatological expertise to model facial skin blemishes. Our method allows for precise, natural, and authentic manipulation of skin blemishes. It is effective across various skin tones and blemishes and does not require large data sets for learning blemish patterns, achieving comparable results to deep learning-based algorithms.

However, CGFR has limitations, such as the need for manual selection of blemishes and untested performance in complex lighting conditions. Despite these challenges, our approach offers new opportunities in the cosmetic industry, with potential for innovation in skin care products. This research contributes valuable insights into the intersection of computer vision and skin science.
\bibliographystyle{IEEEbib}
\bibliography{icme2023template}


\end{document}